\documentclass{article}

\usepackage{microtype}
\usepackage{graphicx}
\usepackage{subfigure}
\usepackage{booktabs} 

\usepackage{hyperref}

\usepackage{amsmath}
\usepackage{amsfonts}

\usepackage{multirow}
\usepackage{bm}
\usepackage{amssymb}

\usepackage[accepted]{icml2021}

\icmltitlerunning{Zoo-Tuning: Adaptive Transfer from a Zoo of Models}

\begin{document}

\twocolumn[
\icmltitle{Zoo-Tuning: Adaptive Transfer from a Zoo of Models
}



\icmlsetsymbol{equal}{*}

\begin{icmlauthorlist}
\icmlauthor{Yang Shu}{equal,tsinghua}
\icmlauthor{Zhi Kou}{equal,tsinghua}
\icmlauthor{Zhangjie Cao}{tsinghua}
\icmlauthor{Jianmin Wang}{tsinghua}
\icmlauthor{Mingsheng Long}{tsinghua}

\end{icmlauthorlist}

\icmlaffiliation{tsinghua}{School of Software, BNRist, Tsinghua University, Beijing 100084, China. E-mail: Yang Shu (shu-y18@mails.tsinghua.edu.cn)}

\icmlcorrespondingauthor{Mingsheng Long}{mingsheng@tsinghua.edu.cn}

\icmlkeywords{Transfer Learning, Fine-Tuning, Model Zoo}

\vskip 0.3in
]



\printAffiliationsAndNotice{\icmlEqualContribution} 

\begin{abstract}

With the development of deep networks on various large-scale datasets, a large zoo of pretrained models are available. When transferring from a model zoo, applying classic single-model based transfer learning methods to each source model suffers from high computational burden and cannot fully utilize the rich knowledge in the zoo. We propose \emph{Zoo-Tuning} to address these challenges, which learns to adaptively transfer the parameters of pretrained models to the target task. With the learnable channel alignment layer and adaptive aggregation layer, Zoo-Tuning \emph{adaptively aggregates channel aligned pretrained parameters} to derive the target model, which promotes knowledge transfer by simultaneously adapting multiple source models to downstream tasks. The adaptive aggregation substantially reduces the computation cost at both training and inference. We further propose lite Zoo-Tuning with the temporal ensemble of batch average gating values to reduce the storage cost at the inference time. We evaluate our approach on a variety of tasks, including reinforcement learning, image classification, and facial landmark detection. Experiment results demonstrate that the proposed adaptive transfer learning approach can transfer knowledge from a zoo of models more effectively and efficiently.

\end{abstract}

\section{Introduction}
Transfer learning leverages knowledge from existing datasets to improve the learning on the target task~\cite{cite:TKDE09TransferLearningSurvey}. With deep learning achieving state-of-the-art performance on various machine learning tasks~\cite{cite:Nature15DeepLearning,cite:BookDeepLearning}, deep transfer learning has attracted more attention in recent years~\cite{tan2018survey}. A standard deep transfer learning paradigm is to leverage models pretrained on large-scale datasets~\cite{cite:IJCV15Imagenet} and \emph{fine-tune} the model on the target task~\cite{cite:NIPS14HowTransferable}, which is demonstrated to be a simple and effective solution in real-world applications  \cite{cite:CVPR19DoBetterTransfer}.

However, most of the existing fine-tuning approaches only transfer from a single pretrained deep network to the target task~\cite{cite:ICML18L2SP,cite:ICLR19Delta,cite:NIPS19BSS}. With the increasing number of large-scale datasets in various fields, we usually have access to a zoo of deep models pretrained on various datasets with different methods such as supervised learning~\cite{cite:CVPR16ResNet}, self-supervised learning~\cite{cite:NIPS14Exemplar} and unsupervised learning \cite{cite:ICML20SimClr}. Therefore, the large diverse model zoo 
defines a new problem setting, \textit{Transfer Learning from a Zoo of Models}, which aims to transfer knowledge from multiple models to promote the learning of the target task.  

The new problem setting introduces new challenges: (1) How to decide the extent of knowledge to transfer from different pretrained models? The different pretrained models contain a diverse range of knowledge. Some pretrained models are more related to the target tasks, while some models are unrelated to or even negatively influence the target learning. For example, although it is common to initialize models with parameters trained on ImageNet, empirical evidence has shown that this practice offers little benefit to applications in medical imaging \cite{cite:NIPS19Transfusion}. Therefore, it is important to decide the correct models to transfer from. (2) How to aggregate the knowledge from different pretrained models? The diverse pretrained models can be complementary to each other, which serve as a more complete knowledge base for the target task. Thus, a model aggregating mechanism is required to integrate knowledge from multiple pretrained models, which further improves the target task. Directly transferring each model to the target and assembling the fine-tuned models may aggregate knowledge but suffers from the large training and inference cost linearly increasing with the size of the model zoo.

In this paper, we propose \textit{Zoo-Tuning}, an effective and efficient solution that enables adaptive transfer from a zoo of models to downstream tasks. We decide to transfer model parameters to avoid the high computation burden of forwarding data through all the pretrained models. We first align channels of different pretrained model parameters by a channel alignment layer since pretrained models are trained on diverse datasets. We then aggregate the pretrained model parameters with the weights controlled by a gating network. The channel alignment layer and the aggregation module are learned, and simultaneously, the source pretrained models are tuned by the target loss signals to adaptively fit the target task. We further develop a lite version of Zoo-Tuning with a unified gating value for all data in the target task computed by the temporal ensemble of average gating values of each data batch. Lite Zoo-Tuning further reduces the computation and storage cost at inference time, which serves as an option to trade-off between efficiency and performance.

The contributions of the paper can be summarized as:

\begin{itemize}
    \item We propose Zoo-Tuning, an adaptive transfer method to enable transfer learning from a zoo of models. Zoo-Tuning aligns channels of source pretrained models and learns data-dependent gating networks to aggregate source model parameters, which transfers knowledge from a zoo of source pretrained models with a low computation budget. All the modules are learned, and simultaneously, the pretrained models are tuned by the target loss signals to fit into the target task adaptively.
    \item We develop a lite version of Zoo-Tuning by the temporal ensemble of average gating values of each data batch, which further saves the inference and storage cost and serves as an optional method to trade-off efficiency and accuracy.
    \item We conduct experiments on a variety of tasks, including reinforcement learning, image classification, and facial landmark detection. Experimental results show that the proposed method outperforms single-model transfer methods (and their extensions) and multiple-model transfer methods while remaining efficient.

\end{itemize}

\section{Related Work}

\noindent \textbf{Transfer Learning.} Transfer learning is a machine learning paradigm to transfer knowledge from source domains to improve the learning of the target task~\cite{cite:TKDE09TransferLearningSurvey}. A promising way to leverage knowledge in pretrained models is to use features extracted by the networks~\cite{cite:CVPR14TransferMidlevel,cite:ICML14Decaf} or fine-tune from pretrained networks~\cite{cite:ECCV14AnalyzingNN,cite:CVPR14RichFeature}. To study the transferability of a pretrained model,~\citet{cite:NIPS14HowTransferable} experimentally quantified the generality versus specificity of neurons in each layer of a deep convolutional neural network. ~\citet{cite:ICML15DAN, cite:TPAMI18DAN} further proposed to learn transferable representations. Recently, many empirical studies consider the effect of transferring from a pretrained model on various downstream tasks and scenarios, such as classification~\cite{cite:CVPR19DoBetterTransfer,cite:Arxiv19VTAB}, few-shot learning~\cite{cite:Arxiv19PretrainedFewShot}, medical imaging~\cite{cite:NIPS19Transfusion}, object detection and instance segmentation~\cite{cite:ICCV19RethinkingPretraining}. Some works utilize knowledge of a pretrained model by regularizing the parameters~\cite{cite:ICML18L2SP} or features~\cite{cite:ICLR19Delta, cite:NIPS19BSS}. Others propose auxiliary tasks to explore category relationships~\cite{cite:NIPS20CoTuning} or intrinsic structures~\cite{cite:Arxiv20BiTuning} of data. However, it remains unclear how to extend single model transfer learning techniques to a zoo of models, where simply assembling every single model is fairly inefficient.

\textbf{Transfer from Multiple Models.} In the face of a zoo of models, recent works seek to predict the transferability of pretrained models to select the best one~\cite{cite:ICCV19NCE,cite:ICIP19Hscore,cite:ICML20LEEP,you2021logme}. These methods suffer from two shortcomings: The selection may be inaccurate, which hurts the performance; Besides, only a single best model is utilized, which wastes other rich knowledge in the whole model zoo. Some works leverage prior knowledge by inserting features from pretrained models~\cite{cite:Arxiv16ProgressiveNet,cite:ICLR19KnowledgeFlow}. These methods need to pass the input data through all models during training or even inference, which may cause high computation and memory costs. Furthermore, these methods use pretrained models to guide the learning of a student model without tuning and adapting the whole zoo to the target. Guidance without adaptation may fail when the downstream tasks are more complex or less similar to the pretrained tasks.

\textbf{Conditional Computation.} Our method is also related to conditional computation~\cite{cite:Arxiv13LowRankConditional,cite:Arxiv14ConditionalComputation}, where parts of the network are active on a per-example basis. \citet{cite:Arxiv13MOE} introduced the idea of using multiple mixture-of-experts \cite{cite:AIReview14MOESurvey} with their own gating networks as parts of a deep model. \citet{cite:Arxiv17OutrageouslyLarge} introduced sparsely-gated mixture-of-experts layers to form outrageously large neural networks. Different from these methods which increase model capacity to absorb sufﬁciently large data by mixing outputs of sub-networks, we aim to transfer knowledge in source models by adaptively aggregating model parameters.

\begin{figure*}[htbp]
    \centering
    \includegraphics[width=.9\textwidth]{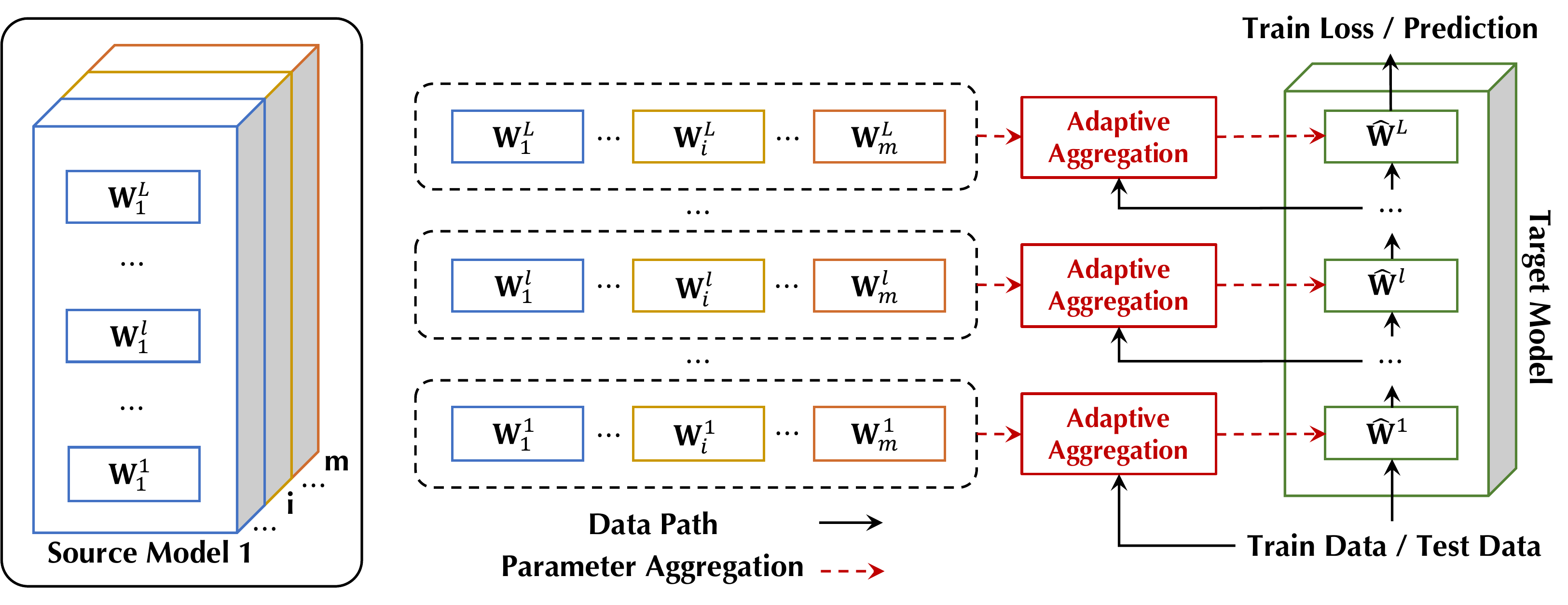}
    \caption{The framework of our proposed method. We derive the target model $\widehat{\mathbf{W}}$ by aggregating the parameters of the source models $\mathbf{W}_i$ in each layer, controlled by the learnable adaptive aggregation modules based on the input data. During training, the adaptive modules are trained, and the source parameters are tuned to transfer to the target task. After training, the tuned source models are aggregated depending on each query data for inference.}
    \label{fig:arc_framework}
\end{figure*}

\section{Approach}
In this section, we introduce the problem of transfer learning from a zoo of models. Then we introduce our Zoo-Tuning approach, with channel alignment and adaptive aggregation. We further propose a more efficient variant, lite Zoo-Tuning. 

\subsection{Transfer Learning from a Zoo of Models}
The most common transfer learning scenario considers only one single pretrained model $M$ at hand to serve the target task $\mathcal{D}=\{(\mathbf{x}, \mathbf{y})\}$. In \textit{transfer learning from a zoo of models}, we consider a more complicated situation where we have a zoo of pretrained models $\mathcal{M}=\{M_1, M_2, \cdots, M_m\}$. This problem is challenging in two ways: (1) The diverse pretrained models hold different relationships to the target tasks, which need transferring knowledge from different pretrained models to different extents; (2) Different models are pretrained on various data and thus store different knowledge, which may be complementary to each other to solve the downstream tasks. How to aggregate knowledge from various pretrained models is an essential but difficult problem for model zoo transfer learning. 

 In this paper, we consider the situation that different models in the zoo have the same architecture but are trained with different \textit{data}, \textit{tasks}, or \textit{pretraining algorithms}. This assumption of the same architecture is reasonable and has its value in practice since architectures such as ResNet~\cite{cite:CVPR16ResNet} can be widely used in various datasets and tasks, and diverse pretrained models of these architectures with rich source knowledge are provided in the open-source community. It is easier and more reliable to apply these models with the same simple and familiar architectures, especially on new problems. Besides, The same architecture enables more effective layer-wise knowledge transfer, which is hard to realize on different architectures. A more generalized situation where models have arbitrary architectures would be interesting and challenging to explore for future work.

\subsection{Zoo-Tuning}
We address the problem of transfer learning from a zoo of models by Zoo-Tuning. The framework is shown in Figure~\ref{fig:arc_framework}. Zoo-Tuning enables knowledge transfer from multiple models by adaptively aggregating source model parameters in each layer, based on the input data, to form the target model. The adaptive aggregation consists of channel alignment and gating networks to control the extent of each source model in transfer learning. As the adaptive aggregation mechanism is lightweight and the target data pass through the derived target model instead of all source models, Zoo-Tuning only introduces similar inference time to a single model, which is computationally efficient. We further propose a lite version of Zoo-Tuning to reduce the storage cost.

\begin{figure*}
    \centering
    \subfigure[Adaptive Aggregation (AdaAgg) Layer]{\includegraphics[width=.48\textwidth]{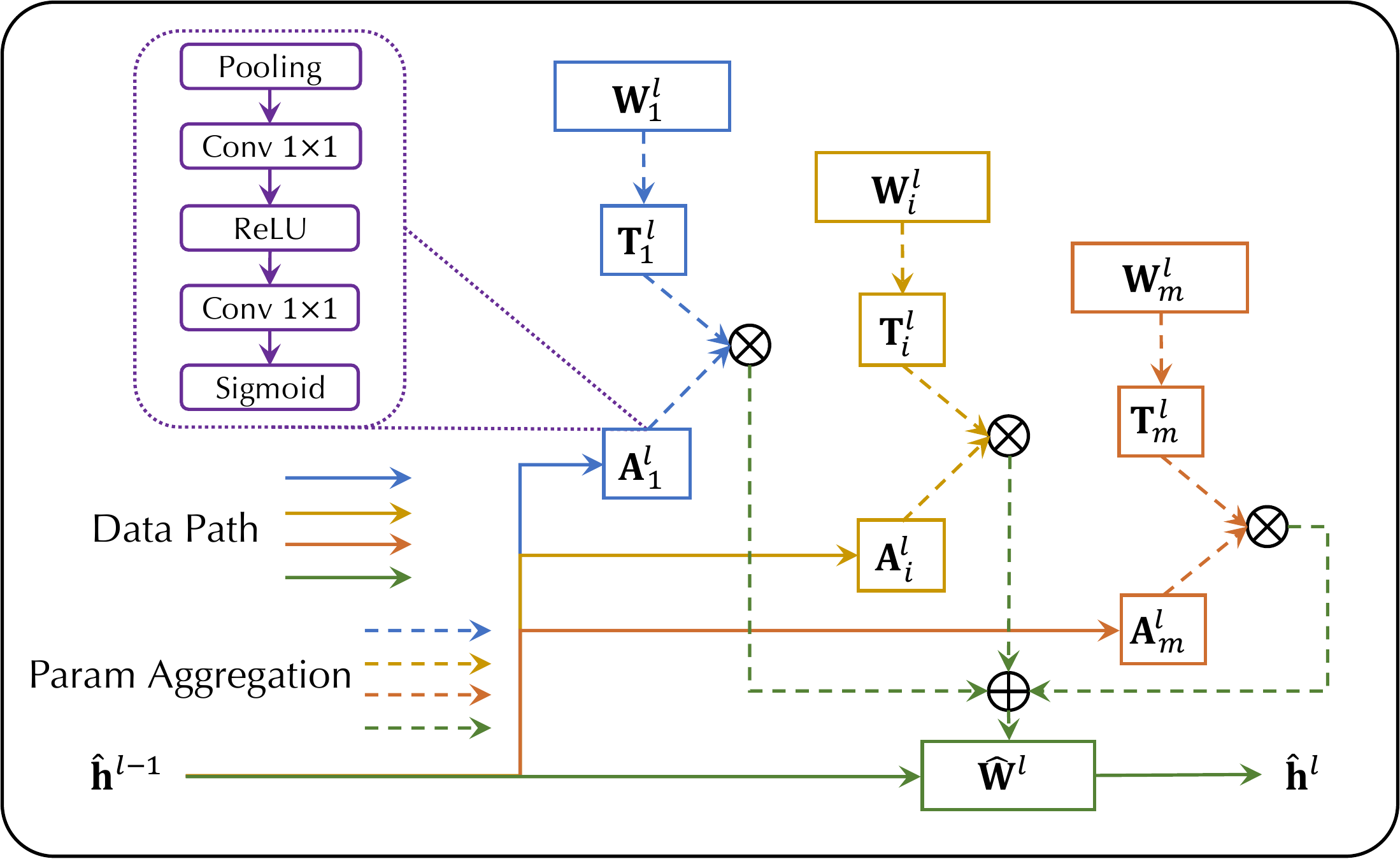}\label{fig:arc2_1}}
    \hfil
    \subfigure[Integrating AdaAgg Layers in a Residual Block]{\includegraphics[width=.44\textwidth]{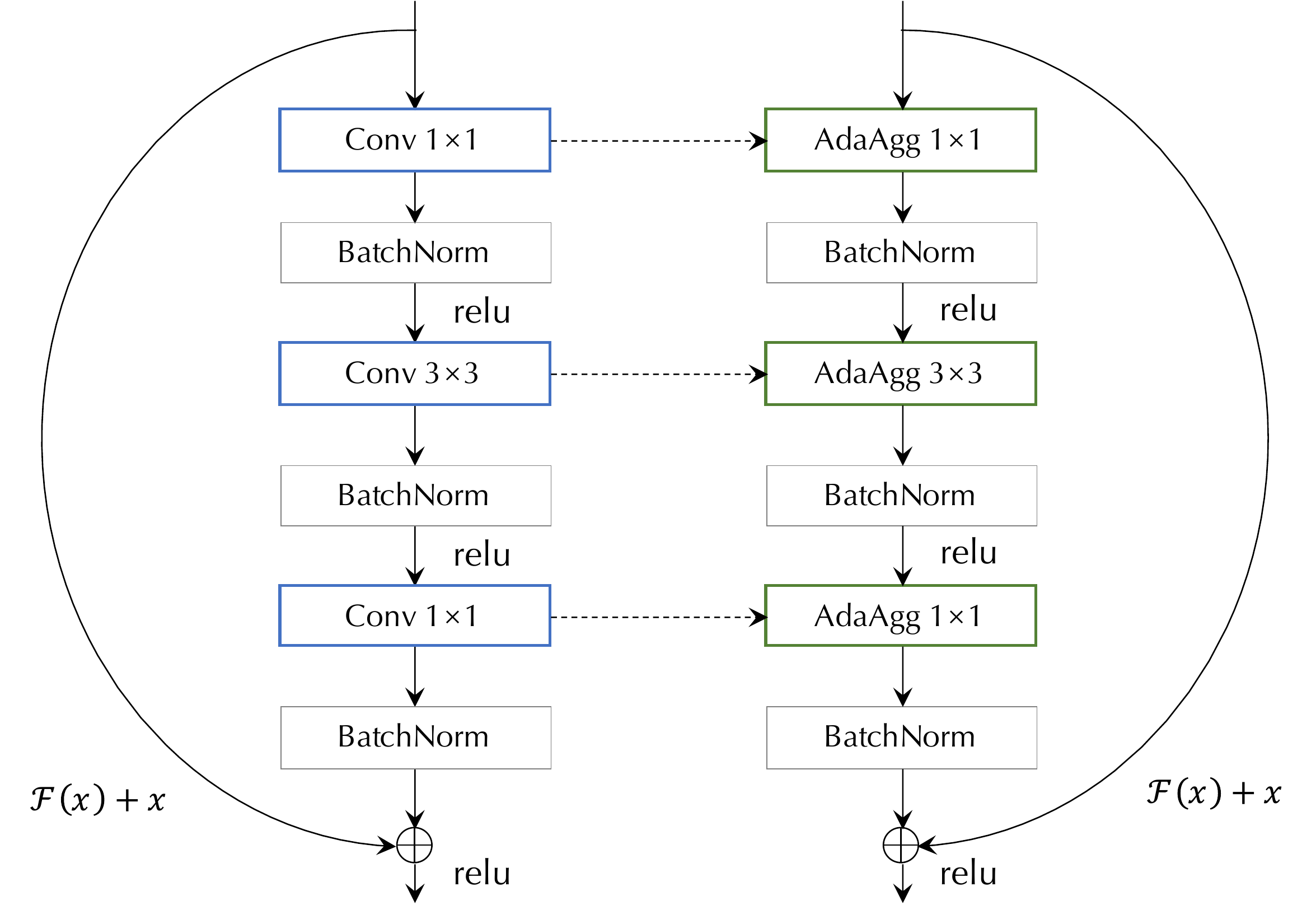}\label{fig:arc2_2}}
    \caption{(a) Illustration of the Adaptive Aggregation Layer. The target input $\hat{\mathbf{h}}^{l-1}$ goes through the gating networks $\mathbf{A}_i^l$ to compute gating values. The source parameters $\mathbf{W}_i^l$ are first aligned by $\mathbf{T}_i^l$ and then aggregated with these gating values to form the target parameters $\widehat{\mathbf{W}}^l$. The input $\hat{\mathbf{h}}^{l-1}$ is finally forwarded through the layer parameterized by $\widehat{\mathbf{W}}^l$. (b) We can change the layers of the network backbone into AdaAgg layers to aggregate models in the zoo. Here is an example where the backbone is composed of residual blocks.}
    \label{fig:arc2}
\end{figure*}

\textbf{Channel Alignment.} Different models are separately trained on diverse datasets or tasks, so even parameters at the same channel of the same layer in different pretrained models may indicate different semantic meanings. The misaligned channels cause difficulty in aggregating parameters of different pretrained models. To address the problem, we adopt a channel alignment module that transforms and aligns channels of different pretrained models. We consider parameters $\mathbf{W}_i^{l}$ of the $l$-th convolutional layer in any source model $M_i$ with the size $C_\text{out} \times C_\text{in} \times K \times K$, where $C_\text{out}$ is the number of output channels, $C_\text{in}$ indicates input channels, and $K$ is the kernel size of the convolutional layer. We adopt a lightweight convolutional layer $\mathbf{T}_i^{l}$ with $1 \times 1$ kernel of size $C_\text{out} \times C_\text{out} \times 1 \times 1$ as the channel alignment module. Specifically, the channels in the source convolutional parameters $\mathbf{W}_i^{l}$ are reorganized by the channel alignment layer to derive the transformed parameters $\widetilde{\mathbf{W}}_i^{l}$ as follows:
\begin{equation}
    \widetilde{\mathbf{W}}_i^{l} =\mathbf{T}_i^{l}*\mathbf{W}_i^{l},
\end{equation}
where we also use $\mathbf{T}_i^{l}$ to denote the parameters of the alignment module. We show an implementation of the channel alignment module for source parameters of convolutional layers here. The idea is easy to extend to other kinds of layers, such as employing a linear alignment layer for fully connected layers. We initialize the channel alignment layer as an identical mapping, which gives the target model a smooth warm-up from the pretrained weights.

\textbf{Adaptive Aggregation.} With channel-aligned source parameters, we develop an adaptive aggregation (\textbf{AdaAgg}) layer to dynamically aggregate source model parameters. We have two key insights in the design of the AdaAgg layer: (1) Each data point of each downstream task should have a different aggregation since each data point holds specific relationships with source tasks; (2) The aggregation should be computationally efficient for a large number of source models. We integrate these two key insights into the design of the AdaAgg layer. As shown in Figure~\ref{fig:arc2_1}, considering the $l$-th layer of the network, the AdaAgg layer is equipped with a gating network $\mathbf{A}_i^l$ for each source model $M_i$, which controls the mixing of its corresponding parameters $\mathbf{W}_i^l$. The gating network $\mathbf{A}_i^l$ takes the feature of the previous layer in the target model $\hat{\mathbf{h}}^{l-1}$ as the input and outputs the gating value $a_i^l$. The aligned source parameters $\widetilde{\mathbf{W}}_i^{l}$ are aggregated with the gating values to derive the parameters of the target model in this layer $\widehat{\mathbf{W}}^l$ as follows: 
\begin{equation}\label{eqn:data_attention_weight}
    \widehat{\mathbf{W}}^l = \sum_{i=1}^{m} a_i^l \widetilde{\mathbf{W}}_i^{l}= \sum_{i=1}^{m}\mathbf{A}_i^l(\hat{\mathbf{h}}^{l-1}) \left(\mathbf{T}_i^{l} *  \mathbf{W}_i^{l} \right),
\end{equation}
We consider lightweight gating networks to reduce the computation and storage cost of the gating network. For example, for a convolutional layer, the gating network consists of a global average pooling layer, $2$ convolutional layers with $1 \times 1$ kernel, and a sigmoid activation function. Such design brings little additional computational cost of the gating network compared to processing data with the original convolution operation, even with a large-scale model zoo.

We can easily change the backbone layers of source models into AdaAgg layers to aggregate source models' parameters in each layer. In Figure~\ref{fig:arc2_2}, we give an example of the residual block. With the target model parameters, the target data are passed through the target model for training and inference. We can solve the optimization problem of adapting the model zoo to the target task as follows:
\begin{equation}
\begin{aligned}\label{eqn:learning_objective}
    &\min_{\mathbf{\Theta}} \mathbb{E}_{(\mathbf{x},\mathbf{y})\sim \mathcal{D}}\mathcal{J}\left(f^L(\cdot; \widehat{\mathbf{W}}^L) \circ \cdots \circ f^1(\mathbf{x}; \widehat{\mathbf{W}}^1), \mathbf{y}\right), \\
\end{aligned}
\end{equation}
where $\mathcal{D}=\{(\mathbf{x}, \mathbf{y})\}$ denotes the target dataset, $L$ is the total number of layers, $f^l$ is the operation of the $l$-th layer parameterized by $\widehat{\mathbf{W}}^l$ and $\mathcal{J}$ is the loss for the target task. $\mathbf{\Theta}=\left( \mathbf{W}_i, \mathbf{A}_i, \mathbf{T}_i \right)$ includes source model parameters $\mathbf{W}_i^l$, channel alignment parameters $\mathbf{T}_i^l$, and gating network parameters $\mathbf{A}_i^l$ in all AdaAgg layers. All of these parameters are adaptively trained or tuned to fit for the target task.

\begin{figure}[ht]
    \centering
    \includegraphics[width=.48\textwidth]{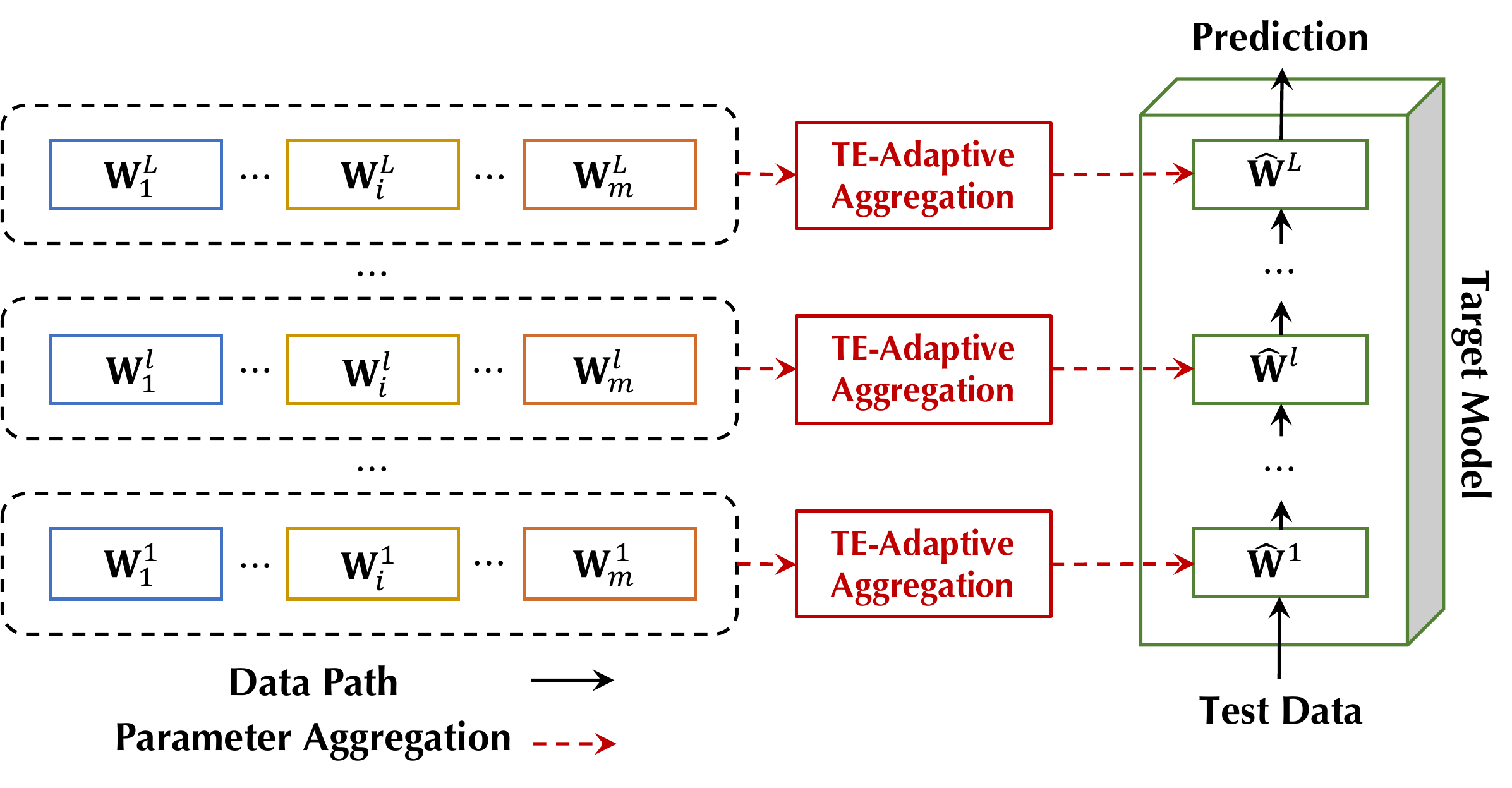}
    \caption{Lite Zoo-Tuning learns a unified inference model for all data in the target task based on the temporal ensemble of adaptive aggregation values over batches. Thus, we just need to store the aggregated target model $\widehat{\mathbf{W}}$ for all target data during inference.}
    \label{fig:arc_lite}
\end{figure}

\subsection{Lite Zoo-Tuning} 
The adaptive aggregation introduced above is computationally efficient but still requires all the source parameters at the inference stage, as the gating values can be computed only when the target data is presented at inference time. As shown in Figure~\ref{fig:arc_lite}, to further save the storage for applying Zoo-Tuning to devices with limited resource, we relax the dependency of the gating values on each individual target sample to the dependency on the entire dataset, by temporal ensemble of the adaptive aggregation over batches. This results in a \textit{unified gating value} and model for all data during inference. During training, for a layer $l$ of a source model $i$, we firstly compute gating values for each sample in the batch, denoted as $a_{i,j}^{l}$, where $j$ is the index of the sample. Then we compute the batch average gating values and their temporal ensemble (TE) over the training batches as follows:
\begin{equation}\label{eqn:task_attention}
\begin{aligned}
\Bar{a}_{i}^{l} = \lambda \cdot \Bar{a}_{i}^{l} + (1 - \lambda) \left(\frac{1}{b}\cdot \sum_{j=1}^{b} a_{i,j}^{l}\right).
\end{aligned}
\end{equation}
We use the batch average values for all training data in the batch and update the temporal ensemble values with $\lambda=0.9$, which is commonly used in temporal ensemble techniques. The temporal ensemble values reflect how the target task relies on each source pretrained model and thus serve as the unified gating values for all target data in inference. Now all target data share the same gating values, so we can pre-compute the aggregation of source parameters to form the target model before inference as follows:
\begin{equation}\label{eqn:task_attention_weight}
    \widehat{\mathbf{W}}^l = \sum_{i=1}^{m} \Bar{a}_{i}^{l} \widetilde{\mathbf{W}}_i^{l}.
\end{equation}
The key difference between Eqn.~\eqref{eqn:data_attention_weight} and  Eqn.~\eqref{eqn:task_attention_weight} is that $\Bar{a}_{i}^{l}$ in Eqn.~\eqref{eqn:task_attention_weight} is shared by all test data and does not change with each sample. So at inference, the target data can be directly forwarded through the pre-aggregated target model. Thus, the cost of lite Zoo-Tuning model in computation and storage is close to a single model. We coin Eqn.~\eqref{eqn:task_attention_weight} Temporal Ensemble Adaptive Aggregation (\textbf{TE-AdaAgg}).

\subsection{Complexity Analysis}\label{sec:complexity}

As we propose layer-wise adaptive transfer of source parameters, for simplicity, we only analyze the complexity of one layer, which can be extended to the whole model. We consider aggregating convolutional layers of $m$ pretrained models. Suppose the dimension of the layer is $C_\text{out} \times C_\text{in} \times K \times K$ where $C_\text{out}$ and $C_\text{in}$ are the numbers of output and input channels, and $K$ is the kernel size. The input feature map has the dimension $C_\text{in} \times H \times W$, where $H$ and $W$ are spatial dimensions. $W$ means the width of the feature map only in this Section~\ref{sec:complexity} to avoid notation abuse. The original convolution operation has the complexity $O \left( H W K ^{2} C_\text{out} C_\text{in} \right)$. Zoo-Tuning introduces additional computation for channel alignment, gating networks, and adaptive aggregation with the cost of $O \left( m K ^{2} C_\text{out}^{2} C_\text{in} \right)$, $O \left( H W C_\text{in} + m C^{2}_\text{in} \right)$ and $O \left( m K ^{2} C_\text{out} C_\text{in} \right)$ respectively, which is $O \left(m K ^{2} C_\text{out}^{2} C_\text{in} + H W C_\text{in} + m C^{2}_\text{in} \right)$ in total. At inference, since the channel alignment is data-independent and can be pre-computed, the computational cost becomes $O\left(m K ^{2} C_\text{out} C_\text{in} + H W C_\text{in} + m C^{2}_\text{in} \right)$. Compared with the original cost of convolution operation, the additional cost for Zoo-Tuning is small. We will also empirically compare the computational complexity of other transfer learning methods and different variants of Zoo-Tuning.

\section{Experiments}

We conduct experiments within three experimental settings. In the first setting, we use a zoo of reinforcement learning models pretrained on various Atari games and transfer to a different set of games. In the other two settings, we use a zoo of diverse computer vision models pretrained on various large-scale datasets and transfer to multiple downstream tasks for classification and facial landmark detection. All experiments are implemented in the PyTorch framework.

\begin{figure*}
    \centering
    \subfigure[Alien]{\includegraphics[width=.32\textwidth]{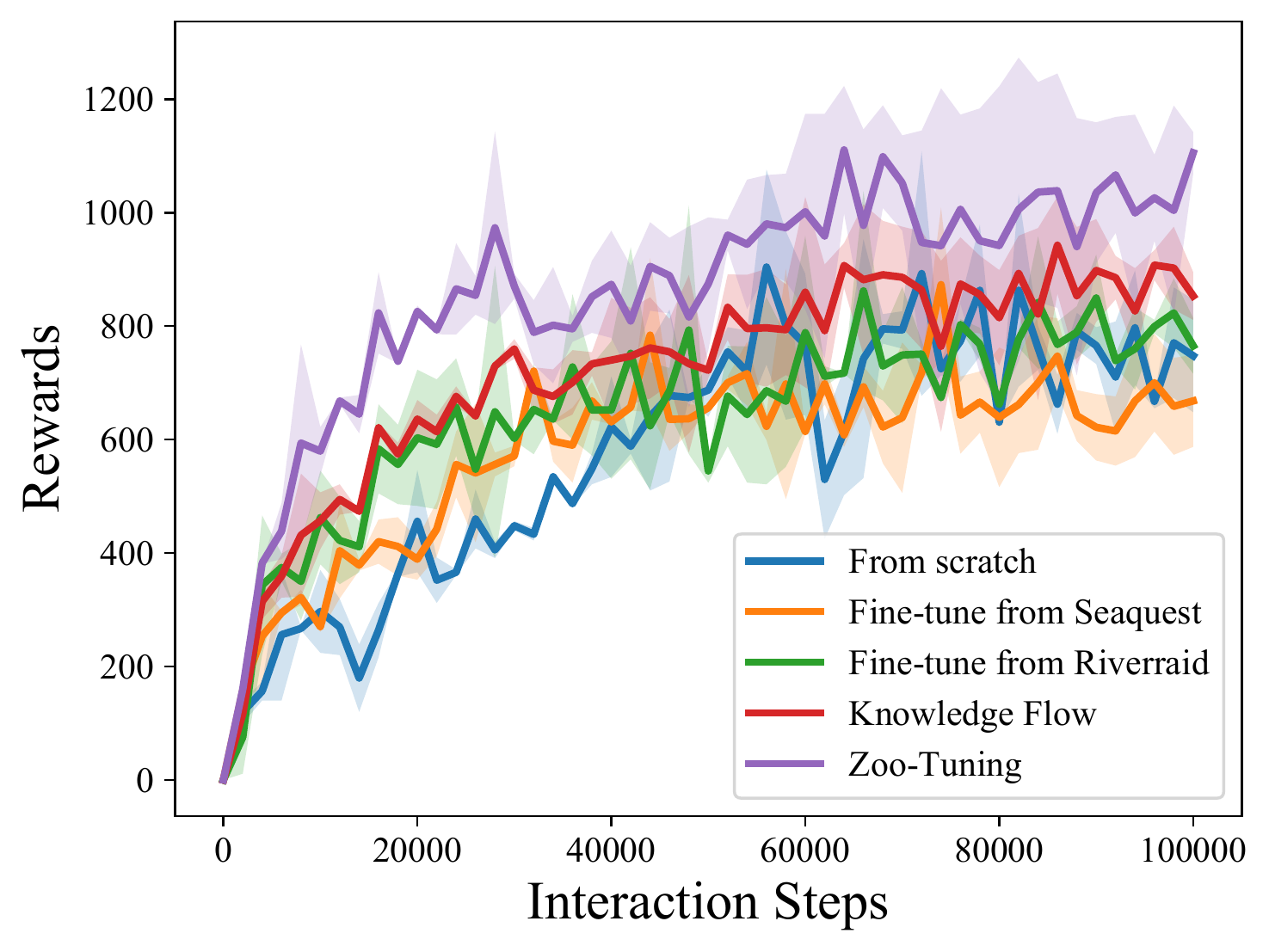}\label{fig:alien}}
    \subfigure[Gopher]{\includegraphics[width=.32\textwidth]{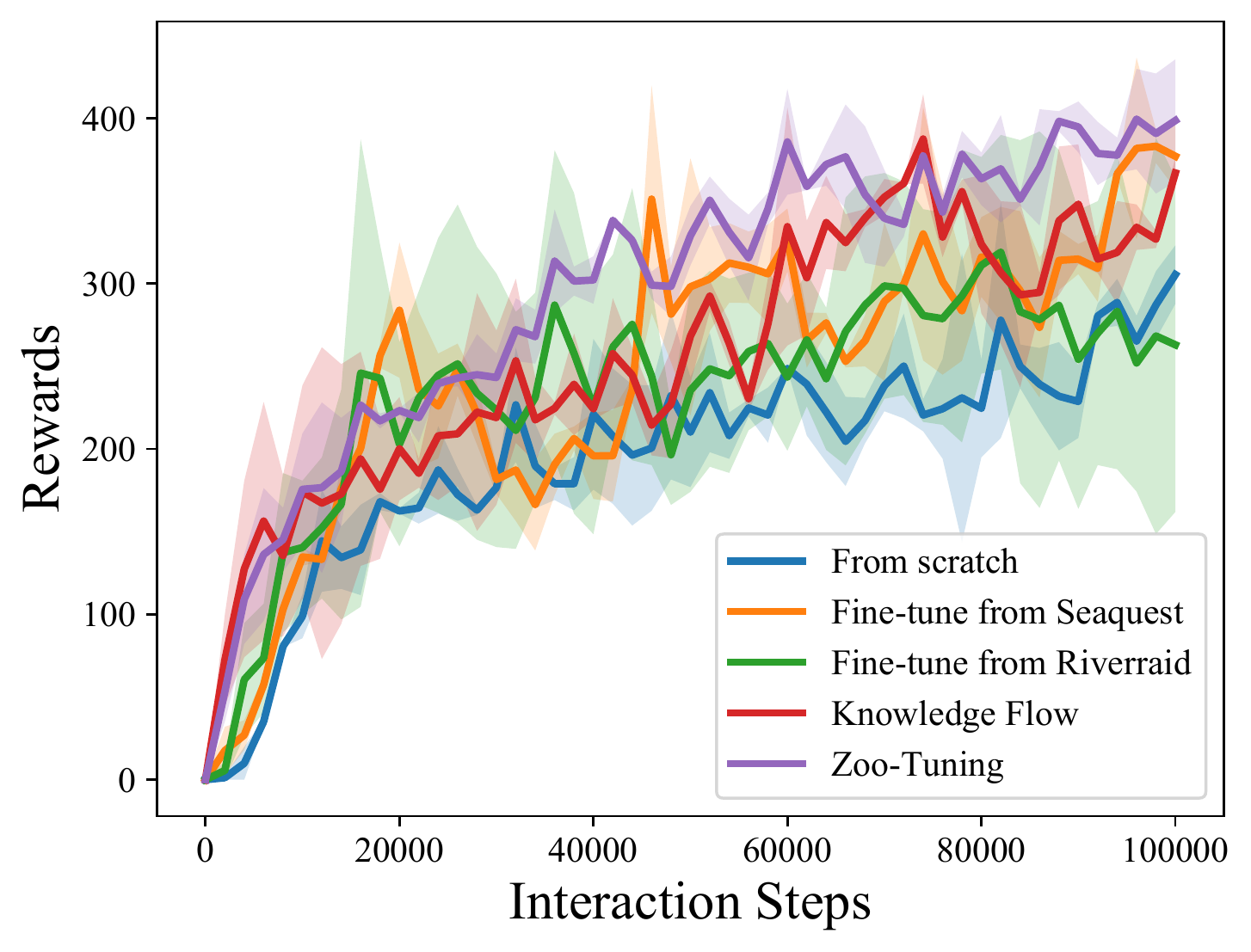}\label{fig:gopher}}
    \subfigure[JamesBond]{\includegraphics[width=.32\textwidth]{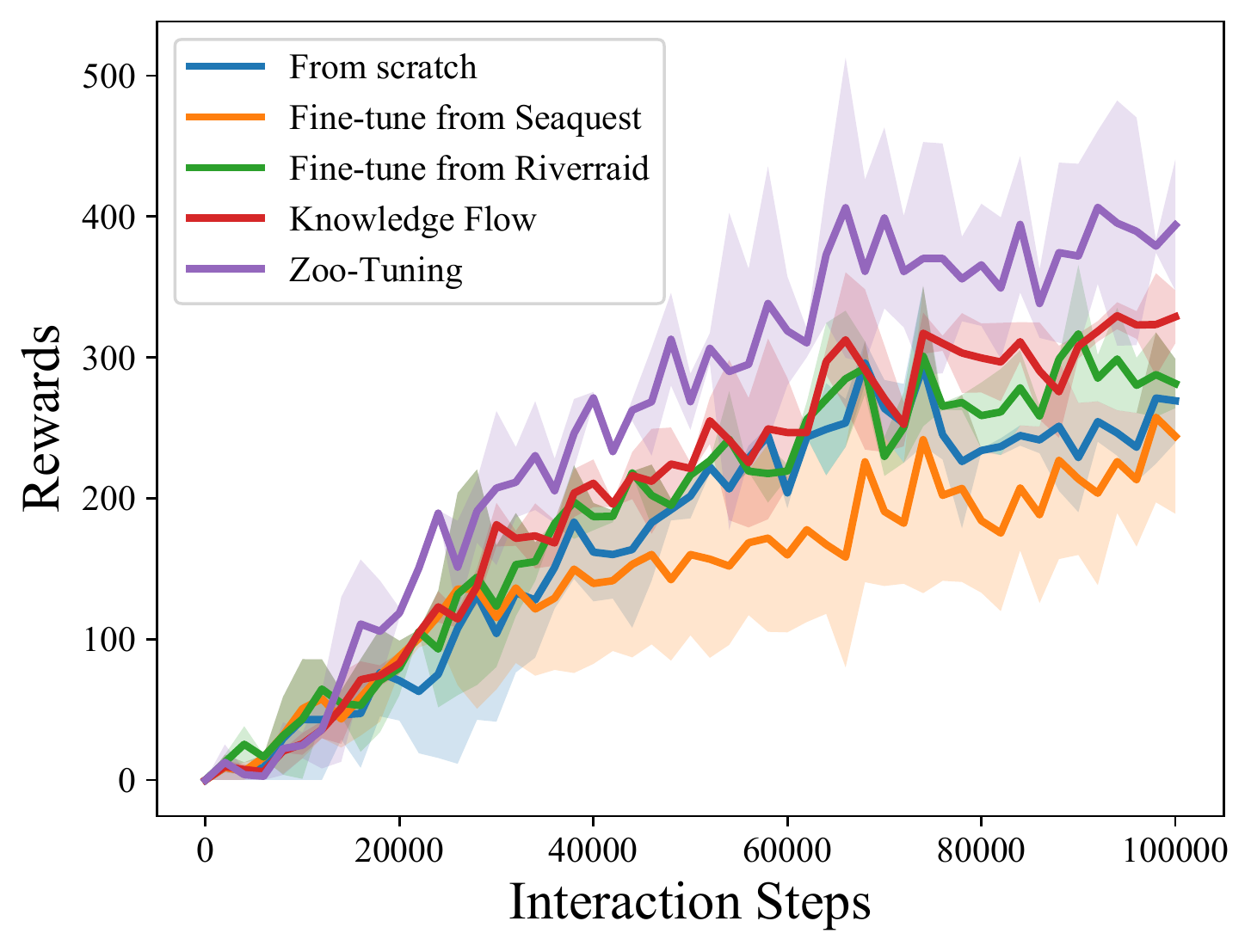}\label{fig:jamesbond}}
    \caption{Results of transferring pretrained models to downstream tasks in the reinforcement learning of Atari games.}
    \label{fig:atari}
\end{figure*}

\subsection{Transfer Learning in Reinforcement Learning}
To demonstrate the generalizability of the proposed Zoo-Tuning method to various domains, we first conduct experiments on reinforcement learning models. 

\textbf{Benchmarks.} Although many reinforcement learning algorithms can achieve much better performance than humans on Atari games, they are still far less efficient than a human learner. So we measure our method at 100k interaction steps (400k environment steps with action repeat of 4) on Atari, which corresponds to the time for a human learner. We use the Seaquest and Riverraid tasks as source tasks and learn an optimal policy by reinforcement learning for each task. We transfer the pretrained reinforcement learning models to $3$ downstream tasks: Alien, Gopher, and JamesBond.

\textbf{Implementation Details.} For the learning algorithm, we follow the implementation of Data-Efficient Rainbow~\cite{van2019use}, which modifies hyper-parameters of Rainbow~\cite{hessel2018rainbow} for data efficiency. The model of Data-Efficient Rainbow consists of $2$ convolutional layers: $32$ filters of size $5\times5$ with the stride of $5$ and $64$ filters of size $5\times5$ with the stride of $5$, followed by a flatten layer and $2$ noisy linear layers~\cite{fortunato2017noisy} with the hidden size of $256$. We use Adam optimizer~\cite{cite:ICLR15Adam} with a learning rate of $1\times10^{-4}$. Other hyper-parameters are kept the same as those in ~\citet{van2019use}.  We repeat each experiment 5 times with different seeds and report the mean and variance of the results.

\textbf{Results.} As shown in Figure~\ref{fig:atari}, in all the downstream tasks, Zoo-Tuning outperforms transferring from a single pretrained model and Knowledge Flow \cite{cite:ICLR19KnowledgeFlow}, which also transfers from multiple pretrained models. The results indicate that Zoo-Tuning enables utilizing knowledge from pretrained policies to promote target tasks by adaptive transfer. Simply fine-tuning from each pretrained model performs similarly with training from scratch, and even slightly worse, demonstrating that when the source and target tasks have a large distribution shift, brutely transferring knowledge may cause negative transfer.

\subsection{Transfer Learning in Image Classification} \label{sec:classification}
For the zoo of models in the classification setting, we use $5$ ResNet-50 models pretrained on representative computer vision datasets: (1) Supervised pretrained model and (2) Unsupervised pretrained model with MOCO~\cite{cite:CVPR20MOCO} on ImageNet~\cite{cite:IJCV15Imagenet}, (3) Mask R-CNN~\cite{he2017mask} model for detection and instance segmentation, (4) DeepLabV3~\cite{chen2018encoder} model for semantic segmentation, and (5) Keypoint R-CNN model for keypoint detection, pretrained on COCO-2017 challenge datasets of each task. In total, the zoo of models are trained on millions of images across a wide range of computer vision tasks, which contains \emph{abundant knowledge in the computer vision domain.} All pretrained models are found in torchvision~\cite{paszke2017automatic} or original implementation.

\begin{table*}[htb]
    \caption{Comparison of top-1 accuracy($\%$) and complexity on the classification benchmarks including the general benchmark, fine-grained benchmark, and specialized benchmark.}
    \label{table:Classification}
    \vskip 0.1in
    \begin{center}
    \begin{sc}
        	\resizebox{0.99\textwidth}{!}{%
            \begin{tabular}{l|ccccccc|c|cccc}
            \toprule
                \multirow{6}*{Model} &  \multicolumn{2}{c}{General} & \multicolumn{3}{c}{Fine-grained} & \multicolumn{2}{c}{Specialized} & & \multicolumn{2}{c}{Train} & \multicolumn{2}{c}{Inference}\\
                & \rotatebox{90}{Cifar-100} & \rotatebox{90}{COCO-70} & \rotatebox{90}{Aircraft} & \rotatebox{90}{Cars} & \rotatebox{90}{Indoors} & \rotatebox{90}{DMLab} & \rotatebox{90}{EuroSAT} & \rotatebox{90}{Avg. Acc.} & \rotatebox{90}{GFLOPs} & \rotatebox{90}{Params} & \rotatebox{90}{GFLOPs} & \rotatebox{90}{Params} \\
            \midrule
                ImageNet sup. & 81.18 & 81.97 & 84.63 & 89.38 & 73.69 & 74.57 & 98.43 & 83.41  &4.12 &23.71M &4.12 &23.71M \\
                MOCO pt. & 75.31 & 75.66 & 83.44 & 85.38 & 70.98 & 75.06& 98.82 & 80.66 &4.12 &23.71M &4.12 &23.71M\\
                MaskRCNN pt. & 79.12 & 81.64 & 84.76 & 87.12 & 73.01 & 74.73 & 98.65 & 82.72 &4.12 &23.71M &4.12 &23.71M\\
                DeepLab pt. & 78.76 & 80.70 & 84.97 & 88.03 & 73.09 & 74.34 & 98.54 & 82.63 &4.12 &23.71M &4.12 &23.71M\\
                Keypoint pt. & 76.38 & 76.53 & 84.43 & 86.52 & 71.35 & 74.58 & 98.34 & 81.16 &4.12 &23.71M &4.12 &23.71M\\
            \midrule
                ensemble & 82.26 & 82.81 &  \textbf{87.02} & \textbf{91.06} & 73.46 & \textbf{76.01} & 98.88 & 84.50 &20.60 &118.55M &20.60 &118.55M\\
                distill & 82.32 & 82.44 & 85.00 & 89.47 & 73.97 & 74.57 & 98.95 & 83.82 & 24.72 &142.28M & 4.12 &23.71M\\
                Knowledge Flow & 81.56 & 81.91 & 85.27 & 89.22 & 73.37 & 75.55 & 97.99 & 83.55 & 28.83 &169.11M & 4.12 &23.71M \\
            \midrule
                Lite Zoo-Tuning & 83.39 & 83.50 & 85.51 & 89.73 & 75.12 & 75.22 & \textbf{99.12} & 84.51 &4.53 &130.43M & 4.12 &23.71M\\
                Zoo-Tuning & \textbf{83.77} & \textbf{84.91} & 86.54 & 90.76 & \textbf{75.39} & 75.64 & \textbf{99.12} & \textbf{85.16} &4.53 &130.43M &4.18 &122.54M\\
                \bottomrule
        \end{tabular}
        }
    \end{sc}
    \end{center}
    \vskip -0.1in
\end{table*}


\textbf{Benchmarks.} We divide the $7$ downstream tasks into three types of benchmarks: general benchmarks, fine-grained benchmarks, and specialized benchmarks, to verify the efficacy of the proposed Zoo-Tuning on different kinds of images. \textbf{(1)} \textit{General} benchmarks with \textbf{CIFAR-100}~\cite{krizhevsky2009learning} and \textbf{COCO-70}: \textbf{CIFAR-100} contains $100$ classes with $600$ images per class, which are split into $500$ training images and $100$ testing images. \textbf{COCO-70} is constructed by cropping objects for each image in COCO dataset~\cite{lin_microsoft_2014} and removing minimal items. It contains $70$ classes with more than $1,000$ images per category. \textbf{(2)} \textit{Fine-grained} benchmarks with \textbf{FGVC Aircraft}~\cite{maji2013fine}, \textbf{Stanford Cars}~\cite{krause_3d_2013} and \textbf{MIT-Indoors}~\cite{quattoni2009recognizing}: \textbf{FGVC Aircraft} is a benchmark for the fine-grained aircraft categorization. It has $100$ categories containing $100$ images each. \textbf{Stanford Cars} contains $16,185$ images for $196$ classes of cars. The data are split into $8,144$ training images and $8,041$ testing images. \textbf{MIT-Indoors} contains $67$ Indoor categories, and a total of $15,620$ images. We use a subset of the dataset that has $80$ images for training and $20$ images for testing per class. \textbf{(3)} \textit{Specialized} benchmarks with  \textbf{DMLab}~\cite{beattie2016deepmind} and \textbf{EuroSAT}~\cite{helber2019eurosat}: \textbf{DMLab} contains frames observed by the agent acting in the DeepMind Lab environment, which are annotated by the distance between the agent and various objects present in the environment. The data are split into $65,550$ training images, $22,628$ validation images and $22,735$ test images. \textbf{EuroSAT} dataset is based on Sentinel-2 satellite images covering $13$ spectral bands and consisting of $10$ classes with $27,000$ labeled and geo-referenced samples.

\textbf{Implementation Details.} We follow the common fine-tuning principle described in~\cite{cite:NIPS14HowTransferable} and replace the last task-specific classification layer with a randomly initialized fully connected layer. We adopt SGD with a learning rate of $0.01$ and momentum of $0.9$ with the same training strategy (total $15$k iterations for fine-tuning with learning rate decay per $6$k iterations) for all pretrained models, compared methods and the proposed Zoo-Tuning. This ensures a fair comparison between different methods and avoids over-tuning on specific tasks. We adopt a batch size of $48$, and all images are randomly resized and cropped to $224\times224$ as the input of the network. More details can be found in supplementary materials.

\textbf{Results.} We report the top-1 accuracy on the test data of each task and the complexity of each method. For our method, we report Zoo-Tuning and lite Zoo-Tuning. For the single-model transfer method, we compare with fine-tuning from every single pretrained model. For methods using all pretrained models, we compare with three methods: using the ensemble of fine-tuned source models for prediction, distilling from the ensemble, and Knowledge Flow \cite{cite:ICLR19KnowledgeFlow}, which is designed to transfer from multiple models. From Table~\ref{table:Classification}. We have the following observations:

On all the three benchmarks, Zoo-Tuning consistently outperforms fine-tuning from each single pretrained model, which indicates that Zoo-Tuning successfully aggregates and utilizes the rich knowledge in the whole zoo of models. 

Compared with the methods using all pretrained models, Zoo-Tuning shows higher or comparable performance on most of the tasks. Compared with the parameters in the model zoo, the additional parameters in Zoo-Tuning is about $10\%$, which shows that the adaptive modules are lightweight. With the adaptive parameter aggregation mechanism, Zoo-Tuning is more computationally efficient. Note that the ensemble predictions require ﬁne-tuning all the candidate pretrained models on the target task firstly. Even at inference time, each query sample should go through all the fine-tuned models to get the ﬁnal prediction, causing high inference cost. Distilling and knowledge ﬂow show similar inference costs as Zoo-Tuning, but Zoo-Tuning achieves higher performance on almost all the tasks. The results demonstrate that Zoo-Tuning is a both effective and efficient solution to transfer learning from a zoo of models. 

Lite Zoo-Tuning also outperforms compared methods in average accuracy. We specially compare it with distilling from the ensemble (Distill) since they are both efficient in inference. Although the performance gain is not large, lite Zoo-Tuning still outperforms Distill consistently on all tasks and achieves greater advantages in the training cost. This is because Distill still needs to forward data through all source models, while lite Zoo-Tuning only needs to pass the data through the aggregated model. Furthermore, Distill needs to fine-tune all the pretrained models on the target data first and then distill a target model from the ensemble outputs of fine-tuned models, which requires a high training cost linearly increasing with the number of source models. The results match the goal of lite Zoo-Tuning to substantially reduce the storage cost in inference while keeping relatively high performance, which is more scalable when training with a large number of source models. 

Zoo-Tuning achieves higher accuracy than lite Zoo-Tuning, which demonstrates that capturing fine-grained data-dependent gating values would help adapt the pretrained models to the target task but with more cost of storage and computation in inference. Lite Zoo-Tuning costs the same GFLOPs and parameters as the single model in inference, with slight performance drop than Zoo-Tuning, which serves as a trade-off between accuracy and efficiency.

\subsection{Transfer Learning in Facial Landmark Detection}

\textbf{Benchmarks.} To explore the usage of Zoo-Tuning on more diverse and complex downstream vision tasks, we use the same model zoo as the image classification tasks in Section~\ref{sec:classification} and consider transferring to three facial landmark detection tasks, \textbf{300W}~\cite{cite:ICCV13300W}, \textbf{WFLW}~\cite{cite:CVPR18WFLW}, and \textbf{COFW}~\cite{cite:ICCV13COFW}. The \textbf{300W} is a combination of HELEN~\cite{cite:ECCV12Helen}, LFPW~\cite{cite:TPAMI13LFPW}, AFW~\cite{cite:CVPR12AFW}, XM2VTS and IBUG datasets, where each face has $68$ landmarks. We follow~\cite{cite:TIP16FaceLocal} and use the $3148$ training images. We evaluate the performance using the full set containing $689$ images. The \textbf{WFLW} dataset is a built on WIDER Face~\cite{cite:CVPR16WiderFace}. There are $7500$ training images and 2500 testing images with 98 manual annotated landmarks. The \textbf{COFW} dataset consists of $1345$ training faces and $507$ testing faces with $29$ facial landmarks.

\begin{table}[ht]
	\caption{Comparison of NME results on facial landmark detection tasks: 300W, WFLW, and COFW.}
	\label{table:Facial}
	\vskip 0.1in
	\begin{center}
	\begin{sc}
				\begin{tabular}{lccc}
					\toprule
					{Model}&{300W}&{WFLW}&{COFW} \\
					\midrule
					scratch & $3.66$ & $5.33$ & $4.20$ \\
					imagenet sup. & $3.52$ & $4.90$ & $3.66$ \\
					moco pt. & $3.45$ & $4.75$ & $3.63$ \\
                    maskrcnn pt. & $3.53$ & $4.87$ & $3.67$ \\
                    deeplab pt. & $3.53$ & $4.89$ & $3.73$ \\
                    keypoint pt. & $3.50$ & $4.90$ & $3.66$ \\
                    \midrule
                    ensemble & $\mathbf{3.33}$ & $4.64$ & $\mathbf{3.46}$ \\
                    distill & $3.45$ & $4.74$ & $3.53$ \\
                    Knowledge Flow & $3.71$ & $5.28$ & $4.58$ \\
				    \midrule
					Zoo-Tuning &$3.41$ & $\mathbf{4.58}$ & $3.51$ \\
					\bottomrule
				\end{tabular}
    \end{sc}
	\end{center}
	\vskip -0.1in
\end{table}

\textbf{Implementation Details.} We generally follow the protocol in~\citet{cite:Arxiv19HRNetFace}. We follow the standard training scheme in~\cite{cite:CVPR18WFLW}. All the faces are cropped by the provided boxes according to the center location and resized to $256\times 256$. We augment the data by $\pm30$ degrees in-plane rotation, $0.75\sim 1.25$ scaling, and random ﬂipping. The models are trained for $60$ epochs with a batch size of $16$. We use Adam optimizer~\cite{cite:ICLR15Adam}. The base learning rate is $1\times10^{-4}$ and is decayed by a rate of $0.1$ at the $30$-th and $50$-th epochs. In testing, each keypoint location is predicted by transforming the highest heat value location to the original image space and adjusting it with a quarter offset in the direction from the highest response to the second highest response~\cite{cite:ICCV17StructureHuman}.

\textbf{Results.} We use the inter-ocular distance as normalization and report the normalized mean error (NME) for evaluation in Table~\ref{table:Facial}. Comparing fine-tuning from each single pretrained model, we can observe that the MOCO~\cite{cite:CVPR20MOCO} pretrained model generally outperforms other pretrained models when transferring to the facial landmark detection tasks. The results confirm that even commonly-used models such as the supervised ImageNet pretrained model cannot dominate all downstream tasks, and it is important to select the more suitable pretrained models for the target task. Zoo-Tuning addresses the challenge by gating networks trained on the target task and consistently outperforms transferring from each single model. Knowledge Flow achieves little improvement than training from scratch and even performs worse on 300W and COFW. This method uses pretrained models as teachers to guide the learning of the student network. But this kind of guidance cannot fully utilize and adapt the knowledge in the zoo, especially when the target tasks are not close to pretrained tasks. Zoo-Tuning adapts the whole model zoo to the target, which is a more effective way of knowledge transfer. Zoo-Tuning achieves comparable performance with the ensemble, but with \textit{much less computational cost} during training and inference.

\begin{figure*}[htbp]
  \centering
  \includegraphics[width=0.96\textwidth]{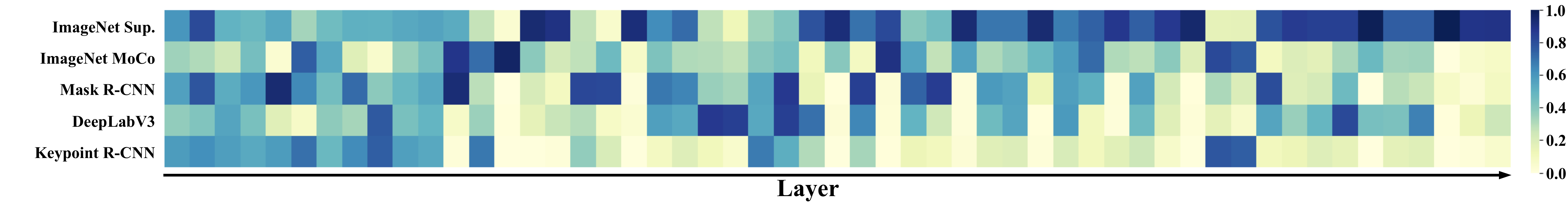}
  \caption{Visualization of gating values in each layer of source models using the five vision pretrained models as source models and using CIFAR-100 as the target task. From the left to right is from the bottom layer to the top layer. The darker color means a higher gating value.}
   \label{fig:vis_weights}
\end{figure*}

\subsection{Analysis}

\noindent \textbf{Variants of Zoo-Tuning.} 
We compare Zoo-Tuning with its variants to demonstrate the efficacy of different modules in Zoo-Tuning. We use all the five pretrained computer vision models described above and transfer them to the COCO-70 dataset and the WFLW dataset. From Table~\ref{table:Ablation}, we have the following observations: (1) Zoo-Tuning outperforms Zoo-Tuning w/o $\mathbf{T}$, which demonstrates that channel alignment of source parameters improves the performance for adaptive aggregation. (2) Zoo-Tuning w/o $\mathbf{T}$ and Zoo-Tuning outperform average aggregation with a large margin. Average aggregation aggregates all the source parameters with the same weight, which treats all source parameters equally. The results demonstrate that it is essential to learn the gating values on the target task and adaptively fit the aggregation to better serve the target task. 

\begin{table}[htbp]
	\caption{Ablation study on variants of Zoo-Tuning.}
	\vskip 0.1in
	\begin{center}
	\begin{sc}
		\label{table:Ablation}
				\begin{tabular}{lcc}
					\toprule
					Method &  COCO & WFLW \\
					\midrule
				    Average Aggregation & 80.53 & 4.75 \\
					Zoo-Tuning w/o $\mathbf{T}$ & 83.92 & 4.64 \\
					Zoo-Tuning & \textbf{84.91} & \textbf{4.58} \\
					\bottomrule
				\end{tabular}
	\end{sc}
	\end{center}
	\vskip -0.1in
\end{table}

\noindent \textbf{Visualization of Gating Values.} We visualize the gating values in each layer of each source model learned by Zoo-Tuning on the CIFAR-100 dataset. As shown in Figure~\ref{fig:vis_weights}, different source models have diverse gating values, which indicates that different source pretrained models have different relationships to the target task, and Zoo-Tuning learns to aggregate the source models for the target task adaptively. Different layers show different gating values, which matches the study on deep networks that different layers contain different knowledge and exhibit different transferability. The gating values show some insights on the transferability of the pretrained models. Though both trained on ImageNet, the supervised model has higher values than the unsupervised one, especially on top layers. Overall, ImageNet supervised model has the highest values, indicating that more related source tasks have more transferable knowledge for the target. Besides, the advantage of the ImageNet supervised model is mainly on top layers, and other networks also have high values in the bottom and intermediate layers. This matches the previous observation that knowledge in deep networks goes from general to task-specific as the layer goes deeper.

\textbf{Number of Parameters.} Zoo-Tuning from a zoo of models outperforms the best single model in the zoo with large margins. However, there is a concern that this improvement may be due to the increase in model parameters. To evaluate how the number of parameters influences the final result, we conduct an ablation study where we change the $5$ diverse pretrained models to $5$ ImageNet supervised pretrained models and perform Zoo-Tuning from the $5$ same models (denoted as $5\times$ ImageNet). We report the results on the image classification datasets COCO-70 and Aircraft. 

\begin{table}[htbp]
	\caption{Ablation study on parameters of the network.}
	\label{table:Parameters}
	\vskip 0.1in
	\begin{center}
	\begin{sc}
				\begin{tabular}{lccc}
					\toprule
					Method &  Aircraft & COCO & Param\\
					\midrule
				    ImageNet & 84.63 & 81.97 & $1\times$ \\
					$5\times$ ImageNet & 84.92 & 82.13 & $5\times$\\
					Lite Zoo-Tuning & 85.51 & 83.50 & $1\times$\\
					Zoo-Tuning & \textbf{86.54} & \textbf{84.91} & $5\times$\\
					\bottomrule
				\end{tabular}
	\end{sc}
	\end{center}
\end{table}

As shown in Table~\ref{table:Parameters}, although the model of transferring from $5\times$ ImageNet has the same parameters as Zoo-Tuning from $5$ diverse pretrained models, it achieves minor improvements compared with fine-tuning from one single model. Note that the proposed lite Zoo-Tuning also outperforms fine-tuning from a single ImageNet model, while lite Zoo-Tuning has the same number of parameters with a single model. The results indicate that the key to the superior performance of Zoo-Tuning is not simply increasing model parameters but adaptively aggregating rich knowledge from the source models. The results not only show the effectiveness of the proposed Zoo-Tuning, but also show the importance of knowledge aggregation in the problem of transfer learning from a zoo of diverse models. 

\textbf{Number of Pretrained Models.} We study how the number of pretrained models in the zoo affects the proposed method. We experiment on the CIFAR-100 dataset by sequentially adding the MoCo, Keypoint R-CNN, DeepLabV3, Mask R-CNN, and ImageNet supervised models into the model zoo. We report the results of Zoo-Tuning with different numbers of pretrained models as well as the results of the single best models in the corresponding zoo in Table~\ref{table:Number}.

\begin{table}[htbp]
	\caption{Ablation study on the number of models.}
	\label{table:Number}
	\vskip 0.1in
	\begin{center}
	\begin{sc}
				\begin{tabular}{lcccc}
					\toprule
					\# of models & 2 & 3 & 4 &5 \\
					\midrule
				    Single Best & 76.38 & 78.76 & 79.12 & 81.18 \\
				    Zoo-Tuning & 78.62 & 79.78 & 81.50 & \textbf{83.77} \\
					\bottomrule
				\end{tabular}
	\end{sc}
	\end{center}
	\vskip -0.1in
\end{table}

From the results, with different model zoo sizes of $2$, $3$, $4$, and $5$ pretrained models, Zoo-Tuning consistently outperforms fine-tuning from the best single model in the zoo, respectively. Note that it is even more difficult to quickly select out the best pretrained model in practice. This indicates that Zoo-Tuning can effectively and efficiently utilize the knowledge in the whole model zoo to promote transfer learning performance, which makes it a better choice in real-world applications. Also, we can find that the performance of Zoo-Tuning increases with the increasing number of pretrained models, which demonstrates that Zoo-Tuning can hold a varying number of pretrained models and is extendable to more source models.

\section{Conclusion}

In this paper, we propose Zoo-Tuning to enable transfer learning from a zoo of models. We align the channels of source parameters with a channel alignment layer and adopt a gating network depending on the input data for each source model to aggregate their parameters, which derives the target model. The channel alignment layer and the gating network are trained, and the source pretrained parameters are tuned by the target task to adaptively transfer knowledge from the zoo of source models to the target task. We further propose lite Zoo-Tuning with the temporal ensemble of batch average gating values, which further reduces the storage cost at inference time. Experiment results in reinforcement learning, image classification, and facial landmark detection demonstrate that Zoo-Tuning achieves state-of-the-art performance with small computational and storage costs.

\section*{Acknowledgments}

This work was supported by the National Key R\&D Program of China (2020AAA0109201), NSFC grants (62022050, 62021002, 61772299), Beijing Nova Program (Z201100006820041), and MOE Innovation Plan of China.

\bibliography{example_paper}
\bibliographystyle{icml2021}

\appendix

\section{Experiment Details}
In this section, we supplement more experiment details for reproducing the experiment results.

\subsection{Implementation Details in Reinforcement Learning}

In the experiments of transfer learning in reinforcement learning, we use the network architecture in Data-Efficient Rainbow, which consists of $2$ convolutional layers: $32$ filters of size $5\times5$ with the stride of $5$ and $64$ filters of size $5\times5$ with the stride of $5$, followed by a flatten layer and $2$ noisy linear layer with the hidden size of $256$. We train two models with this architecture on the Seaquest and Riverraid games, respectively, as our source models. We train for $1\times10^{5}$ steps with the same hyper-parameters as those in Data-Efficient Rainbow. We then perform Zoo-Tuning on the first $2$ layers for each target task. To be specific, we replace the first $2$ layers in the original architecture with adaptive aggregation (AdaAgg) layers, which learn to adaptively aggregate the parameters of these layers in source models to form the corresponding layers of the target model. The last $2$ layers of the architecture are not modified, which are randomly initialized and trained together with the previous AdaAgg layers. During the test stage, we calculate an average of $10$ episodes as the final result, and report the mean and variance of results with 5 different random seeds.

\subsection{Implementation Details in Image Classification and Facial Landmark Detection}

In the experiments of classification and facial landmark detection, we use $5$ pretrained models trained from various datasets and tasks as stated in the main text. Each pretrained model consists of a ResNet-50 backbone and several task-specific head layers. For each target task, we perform Zoo-Tuning on the ResNet-50 backbone. We modify each convolutional layer in the backbone with an AdaAgg layer to aggregate knowledge from all $5$ pretrained models. We do not perform channel alignment on top $1\times 1$ convolutional layers in the backbone, which preserves the accuracy and further reduces the parameters and computational complexity of our method. We keep other layers such as pooling layers, batch normalization layers, and activation layers unchanged. For each batch normalization layer in the backbone, we initialize its learnable linear transformation parameters with the average of the corresponding parameters in $5$ pretrained models, which gives the layer a smooth warm-up. 

Based on the modified backbone with AdaAgg layers, we add a task-specific head for each task. The task-specific head is randomly initialized and trained together with the AdaAgg backbone. For classification tasks, the head is composed of a linear layer. For facial landmark detection, the head is composed of three $4 \times 4$ transpose convolution layers with the stride of $2$, and a $1 \times 1$ convolution layer. For each target task, we run experiments for $3$ times and report the average results of the $3$ runs.

\subsection{Computing Infrastructure}

For reinforcement learning experiments, we implement all the methods based on PyTorch 1.5.0, torchvison 0.6.0, and CUDA 10.2 libraries. For the other two computer vision experiments, we use PyTorch 1.1.0, torchvison 0.3.0, and CUDA 10.0 libraries for the classification tasks. We use PyTorch 1.0.0, torchvison 0.2.1, and CUDA 9.0 libraries for facial landmark detection tasks. We use a machine with 32 CPUs, 256 GB memory, and one NVIDIA TITAN X GPU.

\begin{figure*}[htbp]
  \centering
  \includegraphics[width=.96\textwidth]{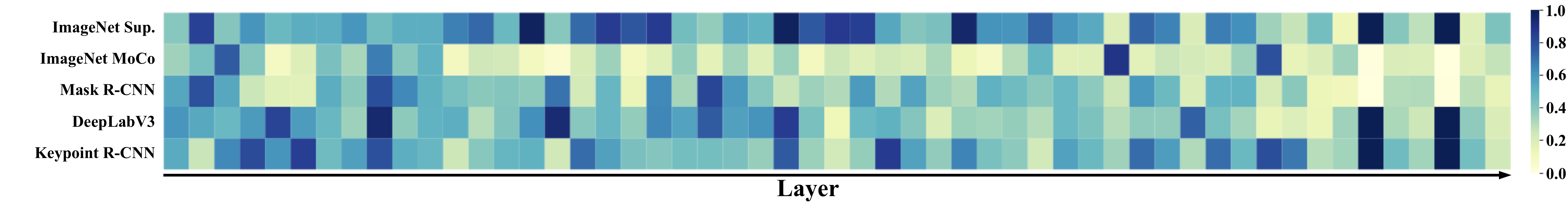}
  \caption{Visualization of gating values in each layer of source models. We use the $5$ vision pretrained models as source models and use the facial landmark detection dataset COFW as the target task. From the left to right is from the bottom layer to the top layer. The darker color means a higher gating value.}
   \label{fig:vis_weights_face}
\end{figure*}

\section{Experiment Results}
In this section, we supplement more experiment results.
\subsection{More Transfer Learning Baselines}
In this section, we design two more transfer learning methods for transferring from a zoo of models. The first method, called Concat-Ensemble-Distill (\textbf{CED}), performs a knowledge distillation from model ensembles but uses the concatenation of their features instead of direct averaging their predictions. Specifically, it adopts a network architecture consisting of a shared backbone with multi-heads to distill the high-level features from each source model in the zoo. Then it uses the concatenation of all these features from multi-heads for the target task. The second method, called Feature Extraction (\textbf{FE}), extracts the features (specifically, features of the penultimate layers) from all the source models in the model zoo and concatenates them to serve as the new features for the target data. It then trains a target classifier with the concatenated features. The two methods are extensions of feature-level transfer learning methods to the model zoo transfer setting. We perform two variants for each method based on whether the source models are firstly fine-tuned with target data (+Finetuned). Note that fine-tuning all source models causes high computation costs.

\begin{table}[ht]
    \caption{Comparison with the proposed baselines.}
	\label{table:Proposed Baselines}
	\vskip 0.1in
	\begin{center}
	\begin{sc}
				\begin{tabular}{lccc}
					\toprule
					{Model}&{Cifar}&{COCO}&{Indoors} \\
					\midrule
				    CED & $81.84$ & $82.19$ & $73.15$ \\
					CED+Finetuned & $81.91$ & $82.01$ & $74.57$ \\
					FE & $80.01$ & $56.52$ & $66.15$ \\
					FE+Finetuned & $83.00$ & $83.34$ & $74.66$ \\
				    \midrule
					Zoo-Tuning &$\mathbf{83.77}$ & $\mathbf{84.91}$ & $\mathbf{75.39}$ \\
					\bottomrule
				\end{tabular}%
    \end{sc}
	\end{center}
\end{table}

As shown in Table~\ref{table:Proposed Baselines}, Zoo-Tuning still outperforms these baselines. Compared with them, Zoo-Tuning has two main advantages. (1) It adaptively leverages the knowledge in pretrained models to facilitate target learning tasks instead of using the fixed features. (2) These baselines only consider high-level features while Zoo-Tuning enables aggregation in all layers, leading to a deeper transfer of knowledge.

\subsection{Visualization of Gating Values}

We also visualize the gating values of each layer in each pretrained model learned by Zoo-Tuning on the facial landmark detection benchmark COFW, as shown in Figure~\ref{fig:vis_weights_face}. 

Different layers in different source models have diverse gating values, which indicates that different layers and source pretrained models have different underlying influences on the target task. Note that this task is more complicated than classification, as the relationship between source and target tasks is unclear, and there is little prior knowledge. We also find that the single model performance cannot necessarily determine its importance when transferring from multiple models under this situation. For example, though fine-tuning from the MoCo pretrained model performs better than other models, it does not have dominant gating values in Zoo-Tuning. This makes transferring from a zoo of models difficult, especially on more complex downstream tasks, and shows the advantage of using Zoo-Tuning, which learns to adaptively aggregate the source models without handcrafted designs. 

\begin{figure}[htbp]
  \centering
  \includegraphics[width=.47\textwidth]{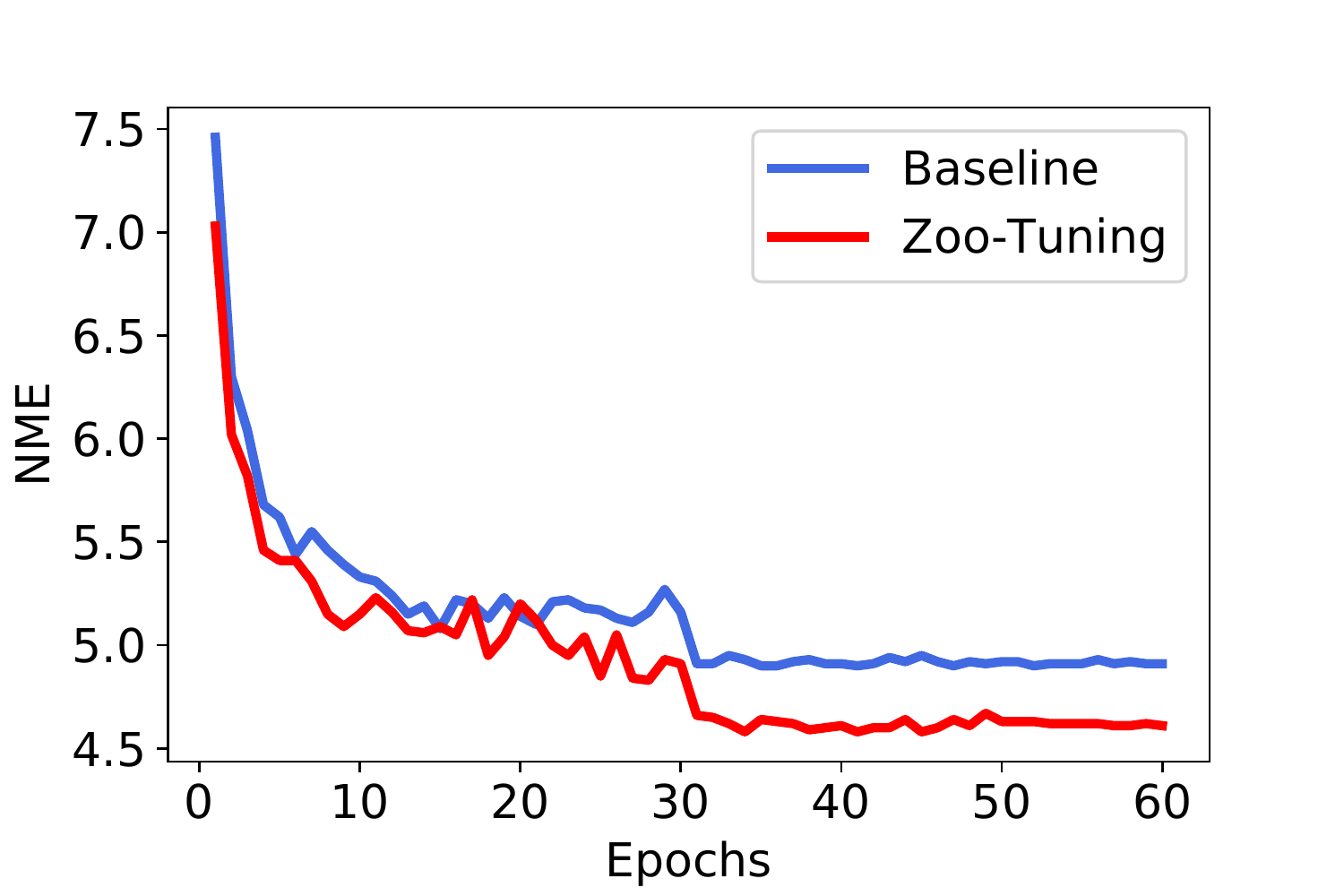}
  \caption{The normalized mean error of the fine-tuning baseline and Zoo-Tuning on WFLW dataset.}
   \label{fig:training_curve}
\end{figure}

\subsection{Training Curves}

We plot the curves of the normalized mean error (NME) on target test data while training with the fine-tuning baseline and Zoo-Tuning methods on the WFLW dataset in Figure~\ref{fig:training_curve}. We can observe that Zoo-Tuning shows a similar convergence speed compared with fine-tuning from one model but consistently achieves better generalization results.

\end{document}